\documentclass[conference]{IEEEtran}
\IEEEoverridecommandlockouts
\usepackage{cite}
\usepackage{amsmath,amssymb,amsfonts}
\usepackage{algorithmic}
\usepackage{graphicx}
\usepackage{textcomp}
\usepackage{xcolor}
\def\BibTeX{{\rm B\kern-.05em{\sc i\kern-.025em b}\kern-.08em
    T\kern-.1667em\lower.7ex\hbox{E}\kern-.125emX}}

\usepackage{amsmath,amssymb,amsxtra,amsthm}
\usepackage[bookmarks=false]{hyperref}

\newcommand{\Bmath}[1]{\mbox{\bf {#1}}}

\newcommand{\BA}{\Bmath{A}}

\newcommand{\BD}{\Bmath{D}}

\newcommand{\Bx}{\Bmath{x}}

\newcommand{\By}{\Bmath{y}}
\newcommand{\Bz}{\Bmath{z}}
\newcommand{\Bw}{\Bmath{w}}

\newcommand{\BW}{\Bmath{W}}

\newcommand{\BV}{\Bmath{V}}

\newcommand{\Bu}{\Bmath{u}}
\newcommand{\Bv}{\Bmath{v}}

\newcommand{\BM}{\Bmath{M}}

\newcommand{\BQ}{\Bmath{Q}}
\newcommand{\Bm}{\Bmath{m}}

\def\W{{\Bmath W}}

\theoremstyle{definition}

\numberwithin{equation}{section}

\usepackage{xcolor}

\usepackage{graphicx}

\usepackage{enumitem}

\usepackage{mathdots}
\usepackage{MnSymbol}

\usepackage{dsfont}

\usepackage{booktabs, multirow}

\title{Predictive Modeling in the Reservoir Kernel Motif Space}

\author{
\IEEEauthorblockN{Peter Ti\v{n}o}
\IEEEauthorblockA{\textit{School of Computer Science} \\
\textit{University of Birmingham}\\
Birmingham, UK \\
p.tino@bham.ac.uk}
\and
\IEEEauthorblockN{Robert Simon Fong}
\IEEEauthorblockA{\textit{Theory Lab, Central Research Institute} \\
\textit{2012 Labs, Huawei Technologies Co. Ltd.}\\
 Hong Kong SAR, China.\\
fong.robert.simon1@huawei.com}
\and
\IEEEauthorblockN{Roberto Fabio Leonarduzzi}
\IEEEauthorblockA{\textit{Theory Lab, Central Research Institute} \\
\textit{2012 Labs, Huawei Technologies Co. Ltd.}\\
 Hong Kong SAR, China.\\
leonarduzzi.roberto.fabio@huawei.com}
}

\begin{document}

\maketitle

\begin{abstract}

This work proposes a time series prediction method based on the kernel view of linear reservoirs. In particular, the time series motifs of the reservoir kernel are used as representational basis on which general readouts are constructed.  We provide a geometric interpretation of our approach shedding light on how our approach is related to the core reservoir models and in what way the two approaches differ.
Empirical experiments then compare predictive performances of our suggested model with those of recent state-of-art transformer based models, as well as the established recurrent network model - LSTM. The experiments are performed on both univariate and multivariate time series and with a variety of prediction horizons.
Rather surprisingly we show that even when linear readout is employed,
 our method has the capacity to outperform  transformer models on univariate time series and attain competitive results on multivariate benchmark datasets. We conclude that simple models with easily controllable capacity but capturing enough memory and subsequence structure can outperform potentially over-complicated deep learning models. This does not mean that reservoir motif based models are preferable to other more complex alternatives - rather,  when introducing a new complex time series model one should  employ as a sanity check simple, but potentially powerful alternatives/baselines such as reservoir models or the models introduced here.

\end{abstract}


\section{Introduction}
\label{sec:intro}






Time series forecasting methods emerge within two main paradigms. The first paradigm essentially "trades time for space/memory," treating the historical time series as a static input to the model, which is then processed spatially. 

Recent successful examples of this paradigm include transformer architectures \cite{vaswani2023attention}. Since their inception, transformers have exhibited exceptional performance in various tasks, including natural language processing (NLP) \cite{devlin2018bert}. Recently, transformers have also been applied to time-series forecasting tasks, as demonstrated in \cite{zhou2021informer, zhou2022fedformer}. However, as \cite{zeng2023transformers} pointed out, whilst multi-head self-attention captures semantic correlations amongst elements of a time series, the temporal correlation is disregarded due to the nature of the static input. 

The second paradigm 
{captures} temporal dependencies in the input data stream through a parametric state-space modelling. Time series data are sequentially encoded into this state space, allowing dynamic capture of temporal information via state-space vectors. Notable examples of this approach include Recurrent Neural Networks and Kalman filters \cite{Kalman1960}.



In this work we consider a special class of Recurrent Neural Networks known as Reservoir Computing models \cite{Jaeger2001,Maass2002,Tino2001,Lukoservicius2009} , where the state space representation and update is fixed and non-trainable. In particular, Echo State Network (ESN)\cite{Jaeger2001,Jaeger2002,jaeger2002a,Jaeger2004} is one of the simplest representations of the Reservoir Computing paradigm, which comprises of a recurrent neural network with a fixed driven dynamical system (reservoir), a fixed input-to-state space mapping, and a simple trainable linear readout map from state space to output.

The term "Reservoir Computing" is derived from the concept that
the reservoir states provide an "echo" of the entire input history, independent of the initial state (Echo State Property). 
To that end, 
the reservoir weights are typically rescaled such that the largest singular value of the state coupling weight matrix is less than $1$.

Echo State Networks (ESNs) have been successfully applied in a diverse range of tasks \cite{Jaeger2004,Bush2005,Tong2007}. Numerous extensions to the classical ESN model have been proposed, including deep ESNs \cite{GALLICCHIO201787}, intrinsic plasticity \cite{Schrauwen2008,Steil2007}, decoupled reservoirs \cite{Xue2007}, leaky-integrator reservoir units \cite{Jaeger2007}, and filter neurons with delay-and-sum readout \cite{holzmann2009}, among others. Furthermore, a similar state-space model approach has been recently rediscovered in \cite{s4} and its associated work.


Recently,  a kernel perspective on linear ESNs was provided in  \cite{tino2020dynamical}, wherein {the state-space representation 
of (potentially left infinite) input time series is viewed as the feature map corresponding to the reservoir kernel. 
Due to the linear nature of the reservoir, the canonical dot product of such feature representations of two input time series can be written analytically as an inner product of the time series themselves. The metric tensor corresponding to this inner product can be used to reveal the representational structure applied by the given  reservoir model to the driving input time series. In particular, the eigenvectors and the associated eigenvalues of the metric tensor correspond to the dominant projection axes (time series `motifs') along with their scaling (`importance') factors, respectively.}



{We propose to use the reservoir time series motifs as the representational basis on which general readouts can be built,
Unlike in the original reservoir model that pre-assigns eigenvalues to motifs, we allow for learnable feature scaling.
%
%
%
Surprisingly, this simple reservoir motif based  architecture outperforms transformers on a number of univariate benchmark problems and achieves comparable performance in in a variety of multi-variate benchmark problems.}

After a brief summary of the kernel view of ESNs in Section \ref{sec:linear_motifs}, we provide a geometric view of our approach in Section \ref{sec:geometry}, where we explain in detail how our approach is related to the core reservoir models and in what way the two approaches differ.
Experiments in section \ref{sec:experiments} compare predictive performances of our suggested model with those of recent state-of-art transformer based models, as well as the established recurrent network model - LSTM. The predictive modelling experiments are performed on both univariate and multivariate time series and with a variety of prediction horizons. Finally, Section \ref{sec:concl} summarizes and discusses the main results of this study.





 


\section{Linear reservoir as a temporal kernel}
\label{sec:linear_motifs}

This section briefly summarizes the kernel view of linear reservoir systems presented in \cite{tino2020dynamical}. Consider a linear reservoir system $R:= (\BW,\mathbf{w},h)$ operating on univariate input, where $\BW \in \mathbb{R}^{N \times N}$ and $\Bmath{w} \in \mathbb{R}^N$ are the dynamic and input  couplings, respectively, and $h:\mathbb{R}^N \to \mathbb{R}^d$ is a trainable readout map. 
%
The corresponding driven linear dynamical system is given by:
\begin{align}
\label{eqn:lrc}
\begin{cases} 
      \Bx(t) = \BW \Bx\left( t - 1\right) +  u(t) \mathbf{w}  \\
      \By(t) = h\left(\Bx\left(t\right)\right),
   \end{cases} 
\end{align}
where $\{u(t)\}_{t} \subset \mathbb{R}$, $\{\Bx(t)\}_{t} \subset \mathbb{R}^N$, and $\{\By(t)\}_{t} \subset \mathbb{R}^d$ denote the {inputs}, {states} and {outputs}, respectively.

Given two sufficiently long time series of length $\tau>N$,
\begin{align*}
\Bu &= \left(u\left(-\tau + 1\right), u\left(-\tau + 2\right),\ldots,u\left(-1\right),u\left(0\right)\right)\\
    &=: \left(u_1,u_2,\ldots,u_\tau\right) \in \mathbb{R}^\tau
\end{align*}
and
\begin{align*}
\Bv &= \left(v\left(-\tau + 1\right), v\left(-\tau + 2\right),\ldots,v\left(-1\right),v\left(0\right)\right)\\
    &=: \left(v_1,v_2,\ldots,v_\tau\right) \in \mathbb{R}^\tau
\end{align*}
we consider the reservoir states reached upon reading them (with zero initial state) their feature space representations \cite{tino2020dynamical}:
\[
\phi(\Bu) = \sum_{j=1}^\tau u_j \BW^{\tau - j}\Bw, \ \ \ 
\phi(\Bv) = \sum_{j=1}^\tau v_j \BW^{\tau - j}\Bw.
\]
The canonical dot product (reservoir kernel)
\[
K(\Bu,\Bv) = \langle \phi(\Bu),\phi(\Bv) \rangle
\]
can be written in the original time series space as a semi-inner product
$\langle \Bu,\Bv \rangle_{Q} = \Bu^\top \BQ \Bv$, where
\[
Q_{i,j} = \Bw^\top \left(\W^\top{}\right)^{i-1} \BW^{j-1} \Bw.
\]

Since the (semi-)metric tensor $\BQ$ is symmetric and positive semi-definite, it admits the following eigen-decomposition:
\begin{align}
\label{eqn:m}
    \BQ = \BM \Lambda_Q \BM^\top ,
\end{align}
where 
$\Lambda_Q := \operatorname{diag}\left(\lambda_1,\lambda_2,\ldots,\lambda_{N_m} \right)$ 
is a diagonal matrix consisting of non-negative eigenvalues of $\BQ$ with the corresponding eigenvectors $\Bm_1,\Bm_2,\ldots,\Bm_{N_m} \in \mathbb{R}^\tau$
(columns of $\BM$). The $N_m := \operatorname{rank}\left(Q\right) \leq N \leq \tau$ eigenvectors of $\BM$ with positive eigenvalues are called the {\em motifs} of $R$. 
We have:
\[
K(\Bu,\Bv) =
\left(\Lambda_Q^{\frac{1}{2}} \BM^\top \Bu\right)^\top \left(\Lambda_Q^{\frac{1}{2}} \BM^\top \Bv\right).
\]

In particular, the reservoir kernel is a canonical dot product 
of time series projected onto the motif space spanned by $\left\{\Bm_i\right\}_{i=1}^{N_m}$:
\[
K(\Bu,\Bv) =
\langle \tilde \Bu, \tilde \Bv \rangle,
\]
where
\begin{align}
\label{eqn:rc_filterview}
\tilde \Bu &= 
\Lambda_Q^{\frac{1}{2}} \BM^\top \Bu \\
&= 
\begin{bmatrix}
     \lambda^{\frac{1}{2}}_1 \cdot \langle \Bm_1, \Bu \rangle \\
     \vdots \\
     \lambda^{\frac{1}{2}}_{N_m} \cdot \langle \Bm_{N_m}, \Bu\rangle 
\end{bmatrix} \\
& =
\left(\lambda_i^{\frac{1}{2}} \cdot \langle \Bm_i,\Bu \rangle \right)_{i=1}^{N_m} \in \mathbb{R}^{N_m}.
\end{align}

\section{Predictive model based on reservoir motifs}
\label{sec:resconv}

In this Section we propose a simple predictive model motivated by the kernel view of Reservoir Computing. In the view of motif space representations
$\tilde \Bu$ of time series $\Bu$ of length $\tau$ ($\tau$-blocks), we propose to also represent the observed $\tau$-blocks of input time series in the reservoir motif space ${span}(\{\Bm_i \}_{i=1}^{N_m})$, but instead of using the associated mixing weights $\lambda_i^{\frac{1}{2}}$ imposed by the reservoir, we allow the {\em motif coefficients}, $C:=\left\{ c_i \in \mathbb{R}\right\}_{i=1}^{N_m}$ to be adaptable:
\begin{align}
\label{eqn:conv_layer}
\varphi\left(\Bu;C\right)
&= 
\begin{bmatrix}
     c_1 \cdot \langle \Bm_1, \Bu \rangle \\
     \vdots \\
     c_{N_m} \cdot \langle \Bm_{N_m}, \Bu\rangle 
\end{bmatrix} \\
& =
\left(c_i \cdot \langle \Bm_i,\Bu \rangle \right)_{i=1}^{N_m} \in \mathbb{R}^{N_m}.
\end{align}

For a real sequence
$\Bz = \{ z(t)\}_{t \in \mathbb{Z}}$, denote by $\Bz(t,\tau) \in \mathbb{R}^\tau$ the $\tau$-block of consecutive elements of $\Bz$ ending with $z(t)$, i.e.
\[
\Bz(t,\tau) = \left(z\left(t-\tau + 1\right),\ldots,z\left(t-2\right), z\left(t-1\right), z\left(t\right)\right).
\]

The $\tau$-block $\Bz(t,\tau)$
will be represented through
\[
          \varphi \left( \Bz\left(t,\tau\right);C\right) := \left(c_i \cdot \langle \Bm_i,\Bz\left(t, \tau\right) \rangle \right)_{i=1}^{N_m}
          \in \mathbb{R}^{N_m},
\]
and based on this representation the associated output will be calculated through a readout map
$q: \mathbb{R}^{N_m} \to \mathbb{R}^d$,
\[
\By(t) = q(\varphi \left( \Bz\left(t,\tau\right);C\right)).
\]

Note that the scaling coefficients $C$ can be actually included in the readout map, so no separate motif weights need to be learnt. Indeed, consider a readout map with unit motif weights
$\tilde{q}:\mathbb{R}^{N_m} \rightarrow \mathbb{R}^d$ defined by:
    \[
        \tilde{q}\left(x_1,\ldots,x_{N_m}\right) := q\left(c_1\cdot x_1,\ldots,c_{N_m} \cdot x_{N_m}\right).
    \]
Then
\[
\tilde{q}\left(\left( \langle \Bm_i,\Bu\left(t, \tau\right) \rangle \right)_{i=1}^{N_m}\right)
        = q\left(\left( c_i \cdot \langle \Bm_i,\Bz\left(t, \tau\right) \rangle \right)_{i=1}^{N_m} \right).
\]

We will thus build predictive models reading out the responses from the recent $\tau$-block histories $\Bz\left(t,\tau\right)$ of input time series $\Bz$ based on orthogonal projections of those histories onto the reservoir motif space (note that the motifs $\{\Bm_i \}_{i=1}^{N_m}$ constitute an orthonormal set of axis vectors):
\begin{align*}
\By(t) &= q(\varphi \left( \Bz\left(t,\tau\right);\mathds{1}\right))\\
&= q(\BM^\top \Bz\left(t,\tau\right)).
\end{align*}

We refer to our model as the {\em Reservoir Motif Machine} (RMM). {Whilst the notion of RMM is inspired by Reservior Computing, it does not belong to the Reservoir Computing paradigm, as illustrated by the next section.}

\section{Geometric view}
\label{sec:geometry}
Given a time series and a sufficiently long $\tau$-block $\Bu \in \mathbb{R}^\tau$ of recently observed items, the reservoir state reached after reading the time series can be approximated by
\[
\phi(\Bu) = \sum_{j=1}^\tau u_j \BW^{\tau - j}\Bw = \BA \Bu,
\]
where $\BA \in \mathbb{R}^{N \times \tau}$ is a linear operator (matrix) with columns $\{ \BW^{\tau - j}\Bw\}_{j=1}^\tau$. 

The operator $\BA$ projects the time series $\Bu$ onto $\mathbb{R}^{N_m}$. 
Consider the SVD-decomposition of $\BA$ given by the eigen-decomposition of 
\[
\BA^\top \BA = \BV \BD \BV^\top,
\]
where 
\[
\BV = [\Bv_1, \Bv_2, ..., \Bv_\tau]
\] 
is a matrix that stores the eigenvectors of $\BA^\top \BA$ as columns and 
$\BD = \operatorname{diag}\left(d_1, d_2,\ldots,d_{\tau} \right)$ 
is a diagonal matrix of the corresponding eigenvalues
$d_1 \ge d_2 \ge \ldots \ge d_{\tau}$. Since 
$d_{N+1} = d_{N+2} = \ldots = d_{\tau} =0$,
geometrically,
the operator $\BA$ maps the unit $\tau$-sphere 
\[
S = \{ \Bx \in \mathbb{R}^\tau | \ \| \Bx\| = 1\}
\]
onto an ellipsoid with semi-axis of length
$d^{\frac{1}{2}}_1,d^{\frac{1}{2}}_2,\ldots,d^{\frac{1}{2}}_{N}$
oriented along 
$\Bv_1,\Bv_2,\ldots,\Bv_{N} \in \mathbb{R}^\tau$.

We now realize that $\BQ = \BA^\top \BA$ and hence {the ellipsoid's semi-axis are equal to the reservoir motifs}. 
The reservoir dynamics maps 
the unit sphere $S$ of possibly long $\tau$-blocks (normalized to norm 1) onto an ellipsoid with semi-axis of length
$\lambda^{\frac{1}{2}}_1,\lambda^{\frac{1}{2}}_2,\ldots,\lambda^{\frac{1}{2}}_{N_m}$
oriented along the reservoir motifs
$\Bm_1,\Bm_2,\ldots,\Bm_{N_m} \in \mathbb{R}^\tau$.
In other words, long histories of input time series are represented through low dimensional representations given projections onto the $N_m \le N < \tau$ dimensional subspace formed the reservoir motif basis.   

We emphasize that our time series representations 
\[
\varphi \left( \Bz\left(t,\tau\right);\mathds{1}\right) =
\BM^\top \Bz\left(t,\tau\right)
\]
are not approximating the reservoir state reached upon reading $\Bz$ up to time $t$, but are projections of $\Bz\left(t,\tau\right)$ onto the feature space defined by the reservoir kernel.

\section{Experiments}
\label{sec:experiments}
In this section we demonstrate the performance of our simple RMM predictive model. To keep the model efficient we constrain ourselves to linear readouts only. We will thus solely employ linear RMMs (Lin-RMM). On a variety of tasks Lin-RMMs are compared with the corresponding reservoir models, as well as with a selection of state-of-the-art methods.

Even though we presented the model development for univariate inputs, the model can be easily extended to the multivariate case e.g., as done here, by representing each of the input univariate channels separately as outlined above and then feeding the joint representation into the linear readout. Other more sophisticated methods based on tensor calculus are possible.

\subsection{Datasets}
To facilitate the comparison of our results with the state-of-the art, we have used
the same datasets and the same experimental protocols used in the recent time series forecasting
papers \cite{zhou2021informer,zhou2022fedformer}. Those are briefly described below for the sake of
completeness. Dataset details are summarized in Table \ref{table:datasets}.

\paragraph{ETT}
The Electricity Transformer Temperature dataset\footnote{https://github.com/zhouhaoyi/ETDataset}
consists of measurements of  oil temperature
and six external power-load features from transformers in two regions of China.
The data was recorded for two years, and
measurements are provided either hourly (indicated by 'h') or every $15$ minutes (indicated by 'm').
The ETTh1, ETTh2 and ETTm1 datasets were used for univariate prediction of the oil temperature
variable,  while the ETTm2 dataset was used for multivariate prediction of all variables.

\paragraph{ECL}
The Electricity Load Diagrams%
\footnote{https://archive.ics.uci.edu/dataset/321/electricityloaddiagrams20112014} consists of
hourly measurements of electricity consumption in kWh for 370 Portuguese clients during four years.
The dataset was used for univariate prediction for client \texttt{MT 320} and multivariate
prediction of all clients.

\paragraph{Weather}
The Local Climatological Data (LCD) dataset%
\footnote{https://www.ncei.noaa.gov/data/local-climatological-data/} consists of hourly measurements
of climatological observations for 1600 weather stations across the US during four years.
The dataset was used for  univariate prediction of the \texttt{Wet Bulb Temperature} variable and for
multivariate prediction of all variables.

\paragraph{Exchange}
The Exchange dataset \cite{lai2018modeling} consists of the daily exchange rates of the currencies of 8
countries (Australia, Great Britain, Canada, China, Japan, New Zealand, Singapore and Switzerland)
from 1990 to 2016.
The dataset was used for multivariate prediction of all exchange rates.

\paragraph{ILI}
The Influenza-Like Illness dataset%
\footnote{https://gis.cdc.gov/grasp/fluview/fluportaldashboard.html}  consists of the weekly data on
the number of patients with influenza-like symptoms, in different age groups, visiting clinics in
the US, between 2002 and 2020.
The dataset was used for multivariate prediction of all variables.


\begin{table}[t]
\centering
\caption{Dataset details. Length, dimension and sampling period for all datasets.}
\label{table:datasets}
\begin{tabular}{cccc}
  \hline\\
  Dataset & Length  & Dimension & Period \\
  \hline\\
  ETTh & 17420 & 7 & 1 hour\\
  ETTm & 69680 & 7 & 15 minutes\\
  ECL & 26304 & 321 & 1 hour \\
  Weather & 52696 & 22 & 1 hour\\
  Exchange &7588 & 8 & 1 day \\
  ILI & 966 & 8 & 1 week \\
  \hline\\
\end{tabular}
\end{table}

\subsection{Experimental setup}
\label{sec:setup}
The typical choice of sampling random reservoir coupling matrices\cite{Jaeger2001,Lukoservicius2009}
requires numerous trials and even luck \cite{Xue2007}, because principled  strategies for the construction of
appropriate random reservoirs  have not been adequately devised.
Alternatively, empirical comparisons and theoretical arguments in \cite{rodan2010minimum,tino2020dynamical} showed that simple cycle reservoirs,
with just  two  degree of freedom - the input weight $r_{in}>0$ and the cycle weight $0<\rho<1$ (which is also spectral radius of the reservoir coupling $\BW$) - can achieve comparable or even superior performance to that of random
reservoirs. Moreover, \cite{li2023simple} showed that such  cyclic reservoirs are universal.
In consequence, we adopted this architecture and constructed the motifs $\{\Bm_i\}$ from Simple Cycle
Reservoirs. The  input weights all have the same absolute value $r_{in}$ and an aperiodic sign pattern generated deterministically based on expansion of an irrational number (in our case $\pi$) 
\cite{rodan2010minimum}. We used grid search on the validation splits to select the spectral radius
$\rho\in\{0.9, 0.99, 0.999, 0.9999\}$ and the input weight $r_{in}\in\{0.01, 0.05, 0.1, 1\}$.
Based on preliminary simulations on the validation sets, we fix  $\tau=336$ for the look-back
window length and reservoir size to $N = 150 < \tau/2 = 168$. The historic horizon $\tau$ is large enough to capture the relevant information for prediction, and
further increasing it showed negligible performance gains. 

To train the linear readout in the Lin-RMM we used ridge regression with ridge
coefficient of $10^{-4}$.


We compare prediction errors obtained with our Lin-RMM with those of state-of-the-art
transformer models published recently in the literature \cite{zhou2021informer,zhou2022fedformer}.
To allow for direct comparisons of the results, we copy the experimental setting of \cite{zhou2021informer,zhou2022fedformer} in that we use:
\begin{itemize}
    \item the \textit{same} datasets with the same  train/validation/test
splits of the data,
    \item the \textit{same} prediction horizons, and
    \item the \textit{same} performance measures, namely mean square and mean absolute error, MSE and MAE, respectively.
\end{itemize}  
The results of the alternative models are reproduced directly from 
\cite{zhou2021informer,zhou2022fedformer}.

For univariate time series, we compare our results with those of the state-of-the-art Informer, a
transformer-based prediction model \cite{zhou2021informer}.
The study \cite{zhou2021informer} reports extensive  comparisons between  Informer and other models such as ARIMA and LSTM, and found the former to outperform all the others.

For multivariate time series, the more recent FEDformer \cite{zhou2022fedformer} was shown to
outperform Informer, so we use the results  reported in \cite{zhou2022fedformer} as a comparison baseline.

\subsection{Time series prediction}

\subsubsection{Univariate prediction}
Table \ref{table:results_multiv} shows the mean-squared and absolute prediction errors, MSE and MAE, respectively, for
univariate time series, compared to the Informer results published in \cite{zhou2021informer},
for several prediction horizons.
First, it can be seen that Lin-RMM surprisingly outperforms all competitors models in both metrics. 

While in principle it may seem surprising that the simple Lin-RMM 
outperforms models with potentially much larger capacity and expressive power, we
conjecture that the explanation lies in the possibility of straightforward inclusion of selective long memory captured in the reservoir motifs and the
relatively low amount of data available for training. While linear models based on motif
convolutions have, in principle, a larger model bias, they can be learned with a much smaller
variance. In consequence, in tasks where the model bias is not too large, the
large variance associated with learning complex models from a small dataset dominates.
As the prediction horizon increases, the tasks become more difficult and the model bias increases;
consequently, the performance gap between the Lin-RMM and Informer decreases.
Still, even in these cases the motif convolution features retain their edge.
More substantial theoretical analysis is the matter of our future work.

\begin{table*}
\centering
\caption{\textbf{Univariate prediction performance. Our Lin-RMM model is compared with the Informer, LSTM and ARIMA models based on their performance results reported in 
\cite{zhou2021informer}
 through the mean square error (MSE) and mean absolute error (MAE).}}
\label{table:results_univ}
\resizebox{0.5\textwidth}{!}{
\begin{tabular}{llrrrrrrrr}
\toprule
        &     & \multicolumn{2}{c}{Lin-RMM} & \multicolumn{2}{c}{Informer} & \multicolumn{2}{c}{LSTMa} & \multicolumn{2}{c}{ARIMA} \\
        &     &   MSE &   MAE &      MSE &   MAE &   MSE &   MAE &   MSE &   MAE \\
\midrule
\multirow{5}{*}{ECL}     &  48 & \textbf{0.155} & \textbf{0.301} & 0.239 & 0.359 & 0.493 & 0.539 & 0.879 & 0.764 \\
                         & 168 & \textbf{0.175} & \textbf{0.322} & 0.447 & 0.503 & 0.723 & 0.655 & 1.032 & 0.833 \\
                         & 336 & \textbf{0.166} & \textbf{0.314} & 0.489 & 0.528 & 1.212 & 0.898 & 1.136 & 0.876 \\
                         & 720 & \textbf{0.164} & \textbf{0.314} & 0.540 & 0.571 & 1.511 & 0.966 & 1.251 & 0.933 \\
                         & 960 & \textbf{0.162} & \textbf{0.312} & 0.582 & 0.608 & 1.545 & 1.006 & 1.370 & 0.982 \\
\cline{1-10}
\multirow{5}{*}{ETTh1}   &  24 & \textbf{0.029} & \textbf{0.127} & 0.098 & 0.247 & 0.114 & 0.272 & 0.108 & 0.284 \\
                         &  48 & \textbf{0.044} & \textbf{0.156} & 0.158 & 0.319 & 0.193 & 0.358 & 0.175 & 0.424 \\
                         & 168 & \textbf{0.079} & \textbf{0.211} & 0.183 & 0.346 & 0.236 & 0.392 & 0.396 & 0.504 \\
                         & 336 & \textbf{0.108} & \textbf{0.254} & 0.222 & 0.387 & 0.590 & 0.698 & 0.468 & 0.593 \\
                         & 720 & \textbf{0.189} & \textbf{0.353} & 0.269 & 0.435 & 0.683 & 0.768 & 0.659 & 0.766 \\
\cline{1-10}
\multirow{5}{*}{ETTh2}   &  24 & \textbf{0.058} & \textbf{0.180} & 0.093 & 0.240 & 0.155 & 0.307 & 3.554 & 0.445 \\
                         &  48 & \textbf{0.083} & \textbf{0.220} & 0.155 & 0.314 & 0.190 & 0.348 & 3.190 & 0.474 \\
                         & 168 & \textbf{0.146} & \textbf{0.298} & 0.232 & 0.389 & 0.385 & 0.514 & 2.800 & 0.595 \\
                         & 336 & \textbf{0.186} & \textbf{0.347} & 0.263 & 0.417 & 0.558 & 0.606 & 2.753 & 0.738 \\
                         & 720 & \textbf{0.275} & \textbf{0.427} & 0.277 & 0.431 & 0.640 & 0.681 & 2.878 & 1.044 \\
\cline{1-10}
\multirow{5}{*}{ETTm1}   &  24 & \textbf{0.010} & \textbf{0.073} & 0.030 & 0.137 & 0.121 & 0.233 & 0.090 & 0.206 \\
                         &  48 & \textbf{0.018} & \textbf{0.098} & 0.069 & 0.203 & 0.305 & 0.411 & 0.179 & 0.306 \\
                         &  96 & \textbf{0.028} & \textbf{0.124} & 0.194 & 0.372 & 0.287 & 0.420 & 0.272 & 0.399 \\
                         & 288 & \textbf{0.053} & \textbf{0.171} & 0.401 & 0.554 & 0.524 & 0.584 & 0.462 & 0.558 \\
                         & 672 & \textbf{0.079} & \textbf{0.209} & 0.512 & 0.644 & 1.064 & 0.873 & 0.639 & 0.697 \\
\cline{1-10}
\multirow{4}{*}{Weather} &  24 & \textbf{0.091} & \textbf{0.208} & 0.117 & 0.251 & 0.131 & 0.254 & 0.219 & 0.355 \\
                         &  48 & \textbf{0.135} & \textbf{0.260} & 0.178 & 0.318 & 0.190 & 0.334 & 0.273 & 0.409 \\
                         & 168 & \textbf{0.222} & \textbf{0.345} & 0.266 & 0.398 & 0.341 & 0.448 & 0.503 & 0.599 \\
                         & 336 & \textbf{0.277} & \textbf{0.391} & 0.297 & 0.416 & 0.456 & 0.554 & 0.728 & 0.730 \\
\bottomrule
\end{tabular}
}
\end{table*}

\subsubsection{Multivariate prediction}
Table \ref{table:results_multiv} shows the mean-squared and absolute prediction errors for
multivariate time series, compared to the FEDformer model (in its Fourier and Wavelet versions, f-FEDformer and w-FEDformer, respectively)
introduced in \cite{zhou2022fedformer}.
It can be seen from Table \ref{table:results_multiv} that, for the potentially more difficult task of
multivariate prediction the performance differences are more balanced: FEDformer offers superior performance in two of the larger datasets (Exchange and Weather), while Lin-RMM dominates in the other two smaller ones (ETTm2, ILI).


\begin{table*}
\centering
\caption{\textbf{Multivariate prediction performance of the studied models.}}
\label{table:results_multiv}
\resizebox{0.5\textwidth}{!}{
\begin{tabular}{llrrrrrr}
\toprule
        &     & \multicolumn{2}{c}{Lin-RMM} & \multicolumn{2}{c}{f-Fedformer} & \multicolumn{2}{c}{w-Fedformer} \\
        &     &   MSE &   MAE &         MSE &   MAE &         MSE &   MAE \\
\midrule
\multirow{4}{*}{ETTm2}    &  96 & \textbf{0.107} & \textbf{0.226} &          0.203 &          0.287 &          0.204 &          0.288 \\
                          & 192 & \textbf{0.140} & \textbf{0.263} &          0.269 &          0.328 &          0.316 &          0.363 \\
                          & 336 & \textbf{0.177} & \textbf{0.302} &          0.325 &          0.366 &          0.359 &          0.387 \\
                          & 720 & \textbf{0.223} & \textbf{0.349} &          0.421 &          0.415 &          0.433 &          0.432 \\
\cline{1-8}
\multirow{4}{*}{Exchange} &  96 &          0.874 &          0.680 &          0.148 &          0.278 & \textbf{0.139} & \textbf{0.276} \\
                          & 192 &          1.857 &          1.025 &          0.271 &          0.380 & \textbf{0.256} & \textbf{0.369} \\
                          & 336 &          2.819 &          1.306 &          0.460 &          0.500 & \textbf{0.426} & \textbf{0.464} \\
                          & 720 &          1.753 &          1.013 &          1.195 &          0.841 & \textbf{1.090} & \textbf{0.800} \\
\cline{1-8}
\multirow{4}{*}{ILI}      &  24 & \textbf{1.549} &          1.005 &          3.338 &          1.260 &          2.203 & \textbf{0.963} \\
                          &  36 & \textbf{1.544} &          1.003 &          2.678 &          1.080 &          2.272 & \textbf{0.976} \\
                          &  48 & \textbf{1.279} & \textbf{0.885} &          2.622 &          1.078 &          2.209 &          0.981 \\
                          &  60 & \textbf{1.119} & \textbf{0.804} &          2.857 &          1.157 &          2.545 &          1.061 \\
\cline{1-8}
\multirow{4}{*}{Weather}  &  96 &          2.677 &          0.876 & \textbf{0.217} & \textbf{0.296} &          0.227 &          0.304 \\
                          & 192 &          3.295 &          0.956 & \textbf{0.276} & \textbf{0.336} &          0.295 &          0.363 \\
                          & 336 &          2.926 &          0.939 & \textbf{0.339} & \textbf{0.380} &          0.381 &          0.416 \\
                          & 720 &          2.373 &          0.912 & \textbf{0.403} & \textbf{0.428} &          0.424 &          0.434 \\
\bottomrule
\end{tabular}
}
\end{table*}

\subsection{Comparisons of RMM with RC}
So far we have demonstrated the surpring power of the Lin-RMM model, but how does it fare when compared to the reservoir model (Simple Cycle Reservoir) from which the motif base was constructed? As we argued in section \ref{sec:geometry} the sequence representations in Lin-RMM are not equal to those in the corresponding reservoir model.

We compare Lin-RMM with the corresponding reservoir models (linear dynamics) with both linear (L-RC) and non-linear (NL-RC) readouts (MLP with cross-validated model complexity).
For the sake
of brevity, we focus only on the results on univariate prediction.

Table~\ref{table:results_rc_comp} shows the prediction errors for all the models under
consideration. It can be seen that performance of all models is roughly similar, except for the
ECL dataset where differences are large.

In general, Lin-RMM generally has a slightly better performance than both RC
models. We interpret this as a confirmation that the versatility that Lin-RMM provides by learning how to
combine the motifs translates in a small performance boost in prediction tasks. Indeed, this
improvement appears to be sufficient enough to compensate for the detrimental effect of the lack of non-linearity.


\begin{table*}
\centering
\caption{\textbf{Comparison of Reservoir models with Lin-RMM.}}
\label{table:results_rc_comp}
\resizebox{0.5\textwidth}{!}{
\begin{tabular}{llrrrrrr}
\toprule
        &     & \multicolumn{2}{c}{Lin-RMM} & \multicolumn{2}{c}{L-RC} & \multicolumn{2}{c}{NL-RC} \\
        &     &   MSE &   MAE &   MSE &   MAE &    MSE &   MAE \\
\midrule
\multirow{5}{*}{ECL}     &  48 & \textbf{0.155} & \textbf{0.301} &          0.228 &          0.369 &          0.341 &          0.451 \\
                         & 168 & \textbf{0.175} & \textbf{0.322} &          0.236 &          0.384 &          0.327 &          0.456 \\
                         & 336 & \textbf{0.166} & \textbf{0.314} &          0.242 &          0.388 &          0.341 &          0.469 \\
                         & 720 & \textbf{0.164} & \textbf{0.314} &          0.249 &          0.392 &          0.330 &          0.462 \\
                         & 960 & \textbf{0.162} & \textbf{0.312} &          0.249 &          0.393 &          0.321 &          0.455 \\
\cline{1-8}
\multirow{5}{*}{ETTh1}   &  24 & \textbf{0.029} & \textbf{0.127} &          0.032 &          0.135 &          0.031 &          0.128 \\
                         &  48 & \textbf{0.044} & \textbf{0.156} &          0.048 &          0.165 &          0.051 &          0.168 \\
                         & 168 & \textbf{0.079} & \textbf{0.211} &          0.091 &          0.226 &          0.109 &          0.249 \\
                         & 336 & \textbf{0.108} & \textbf{0.254} &          0.125 &          0.271 &          0.142 &          0.295 \\
                         & 720 & \textbf{0.189} & \textbf{0.353} &          0.198 &          0.360 &          0.219 &          0.386 \\
\cline{1-8}
\multirow{5}{*}{ETTh2}   &  24 & \textbf{0.058} &          0.180 &          0.072 &          0.199 &          0.058 & \textbf{0.173} \\
                         &  48 &          0.083 &          0.220 &          0.099 &          0.238 & \textbf{0.082} & \textbf{0.212} \\
                         & 168 &          0.146 &          0.298 &          0.164 &          0.313 & \textbf{0.139} & \textbf{0.290} \\
                         & 336 &          0.186 &          0.347 &          0.224 &          0.369 & \textbf{0.185} & \textbf{0.337} \\
                         & 720 &          0.275 &          0.427 &          0.315 &          0.450 & \textbf{0.260} & \textbf{0.406} \\
\cline{1-8}
\multirow{5}{*}{ETTm1}   &  24 & \textbf{0.010} & \textbf{0.073} &          0.012 &          0.078 &          0.011 &          0.074 \\
                         &  48 & \textbf{0.018} & \textbf{0.098} &          0.021 &          0.106 &          0.021 &          0.106 \\
                         &  96 & \textbf{0.028} & \textbf{0.124} &          0.031 &          0.131 &          0.037 &          0.140 \\
                         & 288 & \textbf{0.053} & \textbf{0.171} &          0.054 &          0.173 &          0.083 &          0.216 \\
                         & 672 &          0.079 &          0.209 & \textbf{0.078} & \textbf{0.208} &          0.138 &          0.282 \\
\cline{1-8}
\multirow{4}{*}{Weather} &  24 & \textbf{0.091} & \textbf{0.208} &          0.108 &          0.232 &          0.093 &          0.209 \\
                         &  48 & \textbf{0.135} & \textbf{0.260} &          0.157 &          0.287 &          0.139 &          0.264 \\
                         & 168 & \textbf{0.222} & \textbf{0.345} &          0.259 &          0.380 &          0.223 &          0.353 \\
                         & 336 &          0.277 & \textbf{0.391} &          0.322 &          0.429 & \textbf{0.271} &          0.396 \\
\bottomrule
\end{tabular}
}
\end{table*}



\section{Discussion and conclusion}
\label{sec:concl}

Motivated by the kernel view of reservoir models \cite{tino2020dynamical}, we have introduced a simple, yet potentially surprisingly powerful model structure that decomposes the input signal along the time series motifs extracted from the reservoir kernel. This idea seems to work for tested benchmark data very well even if the overall model structure is kept linear (linear readout).

\begin{figure*}
	\centering	{\includegraphics[width=.8\textwidth]{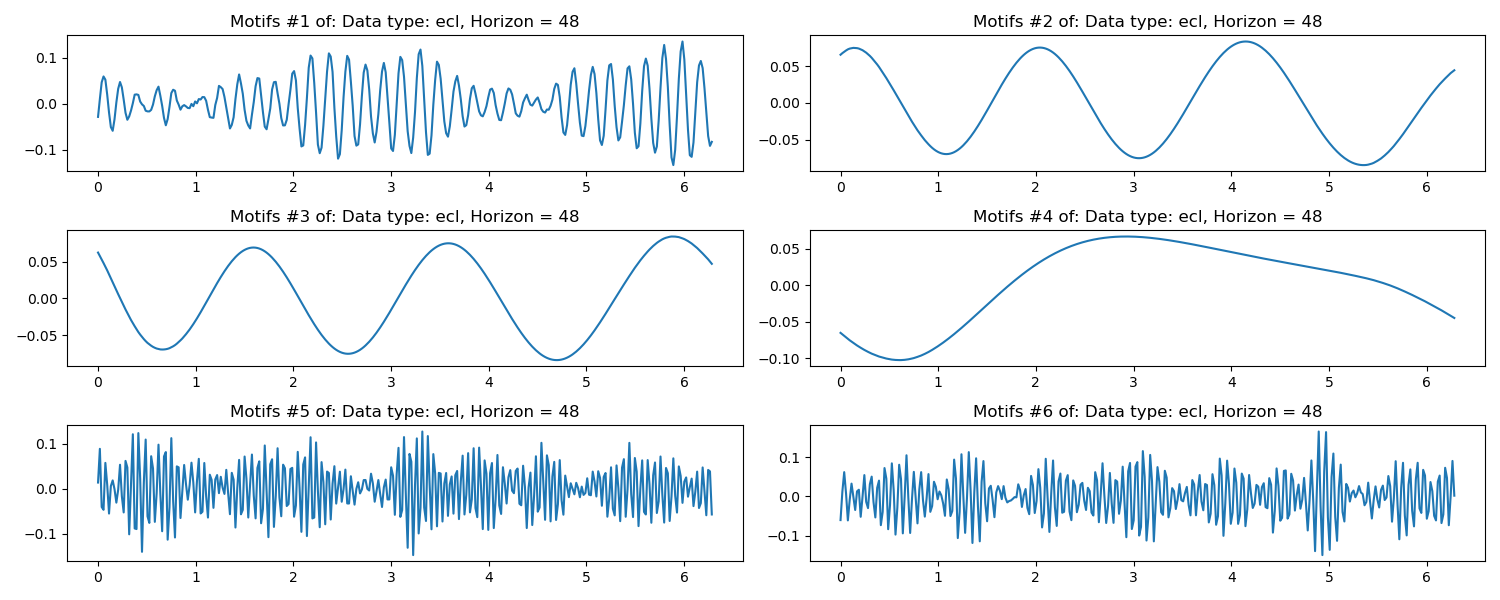}}
	\caption{Six most relevant motifs for the electricity consumption prediction (ECL dataset) - 48h (2 days) prediction horizon.}
	\label{fig:ecl48}
\end{figure*}

\begin{figure*}
	\centering	{\includegraphics[width=.8\textwidth]{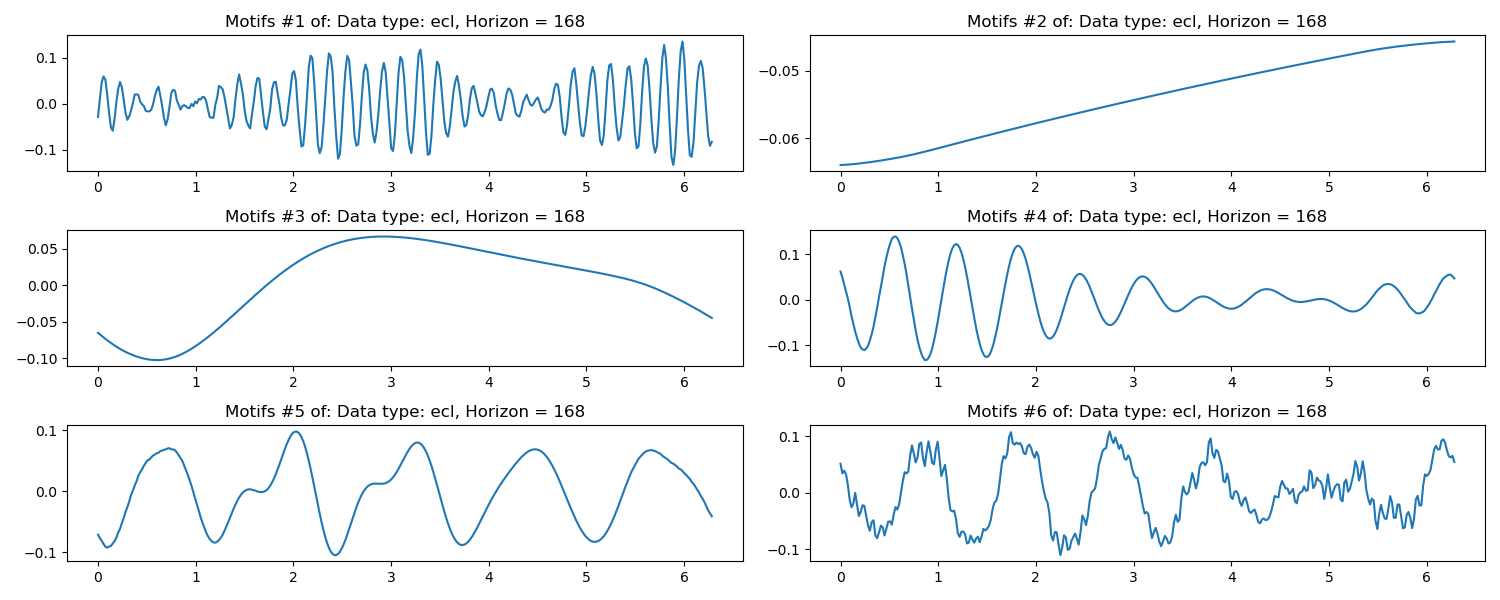}}
	\caption{Six most relevant motifs for the electricity consumption prediction (ECL dataset) - 168h (7 days) prediction horizon.}
	\label{fig:ecl168}
\end{figure*}

Our model is transparent and amenable to analysis. Since the readout map is linear, one can study the magnitudes (absolute values) of the readout weights associated with different motif axis (columns of motif matrix 
$\BM$). As an example, we show in figures \ref{fig:ecl48} and
\ref{fig:ecl168} the six most relevant motifs for the electricity consumption prediction (ECL dataset) associated with the 2 and 7 day prediction horizons, respectively. Interestingly (and intuitively enough), we can observe how the prediction model for the shorter prediction horizon (2 days) concentrates on higher frequency motifs than the model build for the longer 7-day horizon.

\begin{figure*}[b]
	\centering	{\includegraphics[width=.8\textwidth]{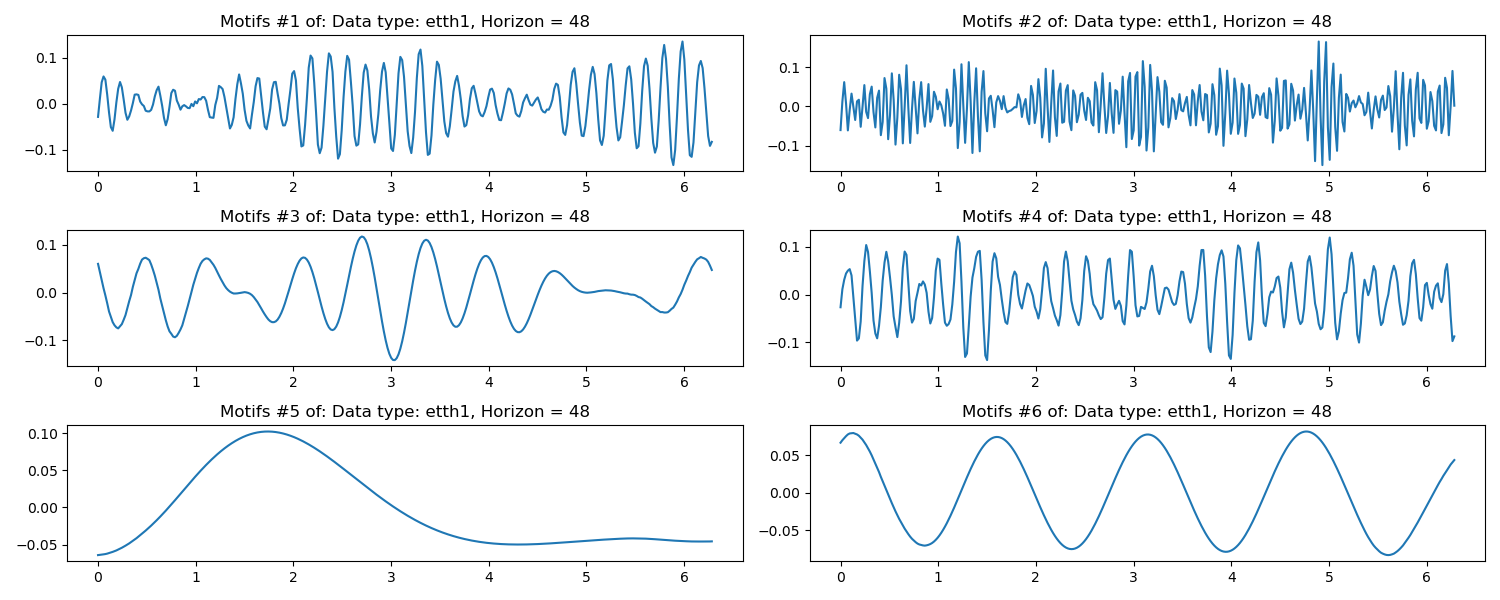}}
	\caption{Six most relevant motifs for the electricity transformer oil temperature prediction (ETTh dataset) - 48h (2 days) prediction horizon, region 1.}
	\label{fig:etth148}
\end{figure*}

\begin{figure*}[b]
	\centering	{\includegraphics[width=.8\textwidth]{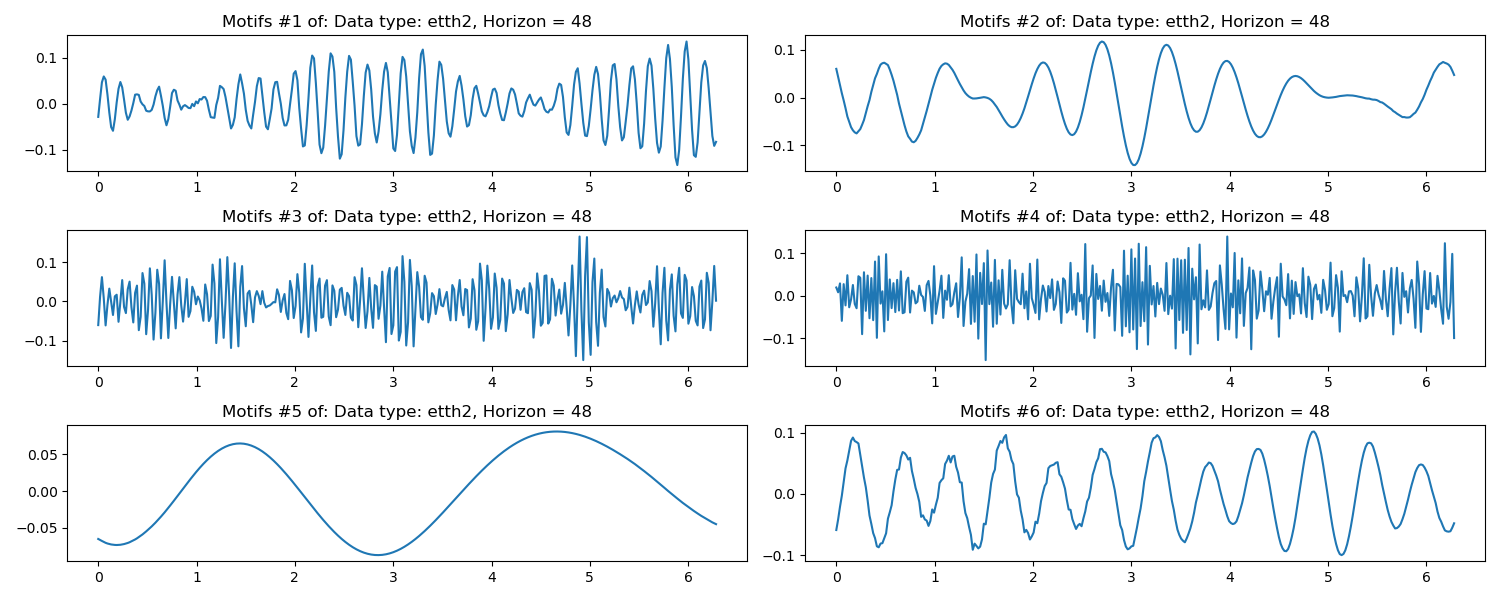}}
	\caption{Six most relevant motifs for the electricity transformer oil temperature prediction (ETTh dataset) - 48h (2 days) prediction horizon, region 2.}
	\label{fig:etth248}
\end{figure*}

As another example, we we present in figures \ref{fig:etth148} and 
\ref{fig:etth248}
six most relevant motifs for the electricity transformer oil temperature prediction (ETTh dataset) employed by RMMs trained on 2-day prediction horizon task using data from two different regions in China. Note how a model for region 2 contains a very high frequency motif missing form the model specializing on region 1. Also a uni-modal "trend" motif for the region-1 model is replaced by the higher frequency bimodal one in the region-2 model. A deeper analysis of this form can help to unveil specific regional differences in the signal characteristics relevant for the prediction task with possible implications for the deployment strategy of electricity transformers

In general, our results demonstrate that simple models with easily controllable capacity that capture enough memory and subsequence structure can outperform (sometimes by a large margin!) potentially over-complicated and over-parametrized deep learning models. Our conclusion is of course not that the reservoir motif based models are preferable to other potentially more complex alternatives, but rather that the temporal structure captured by the reservoir motifs generated from a reservoir model with just two free parameters can be sufficient for excellent predictive performance.
Indeed, this echos the message of \cite{zeng2023transformers}.

We would suggest that when introducing new complex time series models one should always employ simple, but potentially powerful alternatives/baselines such as reservoir models or Lin-RMM introduced here.

%
%
%


\end{document}